\documentclass[letterpaper, 10pt, conference]{ieeeconf}
\IEEEoverridecommandlockouts
\usepackage{graphicx}
\usepackage{amsmath,amssymb}
\usepackage{booktabs}
\usepackage[hidelinks]{hyperref}
\usepackage{xcolor}
\usepackage{tabularx}
\definecolor{oapadd}{RGB}{0,110,70}
\usepackage{tikz}
\usetikzlibrary{arrows.meta,fit,positioning}
\hypersetup{pdftitle={Objectives as Policies: Vision-Language Model Cost Programs for Contact-Rich Manipulation},pdfauthor={Zihao Lu, Trong Gia Hung Nguyen, Fuquan Wang, Yifan Zhu},pdfsubject={Robotics},pdfkeywords={robotic manipulation, vision-language models, model-predictive control}}
\title{\LARGE \bf Objectives as Policies: Vision-Language Model 
Cost Programs for Contact-Rich Manipulation}
\newcommand{\brand}{OaP}

\author{Zihao Lu$^{*}$, Trong Gia Hung Nguyen$^{*}$, Fuquan Wang, and Yifan Zhu%
\thanks{$^{*}$Equal contribution.}%
\thanks{All authors are with the University of Illinois Chicago, Chicago, IL, USA. Contact: \href{mailto:zlu31@uic.edu}{\texttt{zlu31@uic.edu}} and \href{mailto:tnguy321@uic.edu}{\texttt{tnguy321@uic.edu}}.}%
\thanks{This work has been submitted to the IEEE for possible publication. Copyright may be transferred without notice, after which this version may no longer be accessible.}%
}

\begin{document}
\maketitle

\begin{abstract}
Vision-language models (VLMs) have shown strong semantic and spatial reasoning capabilities in 3D environments. Recently, connecting VLMs with model-predictive controllers (MPC) has emerged as a paradigm for combining strong semantic reasoning with physics-based planning. However, connecting these two for contact-rich robotic manipulation requires a task representation that bridges the two and captures how physical interactions can be reasoned by VLMs. In this work, we
introduce Objectives as Policies (\brand{}), a framework that connects a
pretrained VLM to a shared sampling-based MPC through
an editable objective program. Each program specifies a sequence of stages for completing a task, combining running and terminal costs from a cost library with
stage completion conditions. Given a natural-language task instruction and scene observation, we build a simulation-based digital twin through real-to-sim, and the VLM generates an objective program that is executed by the MPC in the simulation. The program is iteratively updated by the VLM based on MPC rollouts, before being deployed in the real world. We evaluate our framework in both simulation and the real world on four tabletop manipulation tasks, and show that it outperforms strong baselines. Additional videos and examples are available on the project website: \url{https://zihaolu001.github.io/oap-project/}
\end{abstract}

\section{INTRODUCTION}
Recently, contact-implicit model-predictive control (MPC) has enabled robots to achieve real-time dexterous manipulations on various tasks~\cite{li2025drop,aydinoglu2024consensus}. Compared to end-to-end learned policies, these physics-based methods naturally generalize to new objects and scenes through online re-planning and ensure safety by directly embedding safety constraints during planning. However, they require careful engineering of the rewards and constraints with the tasks and the scenes known a priori, making deployment into the open world challenging. Meanwhile, end-to-end learned policies, such as vision-language-action models, leverage the semantic understanding of vision-language models (VLMs) trained on internet-scale vision-language data and finetune with robot data to acquire robotic manipulation skills. For example, OpenVLA directly predicts actions from visual observations and language instructions, without an explicit physics model for online planning~\cite{kim2025openvla}. We observe that manipulation skills typically can be decomposed into two key components: \emph{what the robot should be doing, and how the robot can execute it}, where the former requires semantic understanding of the environment and task, and the latter necessitates understanding of the physics and contact dynamics. Hence, in this paper, we ask the following research question: \emph{How can we integrate the semantic understanding of VLMs with physics-based MPC to achieve open-world manipulation skills, without any training data?}

\begin{figure}[t]
\centering
\includegraphics[trim=0 7.2cm 27.3cm 0, clip, width=\columnwidth]{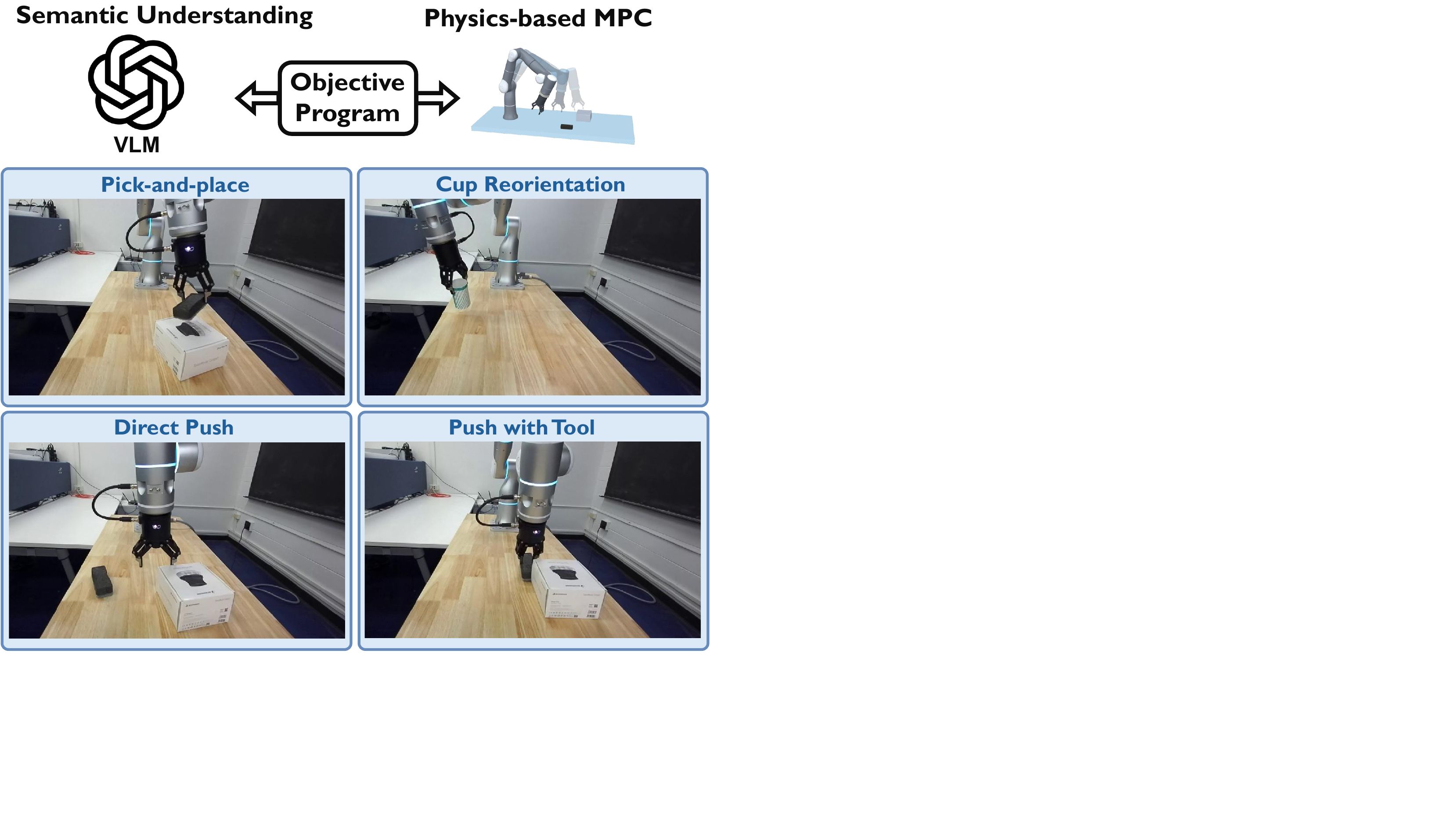}
\caption{Our method combines the semantic understanding of modern vision-language models and physics-based model-predictive control that encodes physical understanding. The two are interfaced through a structured objective program for multi-stage tasks. We show four real-world task executions at the bottom.}
\label{fig:star_figure}
\end{figure}

Previous methods have investigated the use of VLMs for robot manipulation in different ways. VoxPoser and ReKep use 3D value maps and relational keypoint constraints, respectively, for closed-loop planning~\cite{huang2023voxposer,huang2024rekep}. ReKep's local forward model assumes grasped keypoints move rigidly with the end-effector~\cite{huang2024rekep}, whereas our MPC explicitly simulates robot-object contact dynamics. Another method uses VLMs only for estimating the physical parameters of a world model and employs large-language models (LLMs) that script the MPC and go through iterative real-world refinement~\cite{coral2026}. Beyond using these foundation models to generate objectives and constraints for MPCs, some works use VLMs to directly propose robot end-effector paths~\cite{zhao2024vlmpc,zhou2025robotracer}. Another class of methods trains VLMs to predict execution progress~\cite{liang2026robometer}, which can be used as a noisy reward function during offline reinforcement learning training. Different from these methods, we propose a framework where we enhance a VLM with a digital twin obtained through real-to-sim and a library of objective terms, and allow the VLM to iteratively refine a composite objective function from simulated MPC rollouts, before real-world execution. 

\begin{figure*}[t]
\centering
\includegraphics[trim=0 14.7cm 8.9cm 0, clip,width=\textwidth]{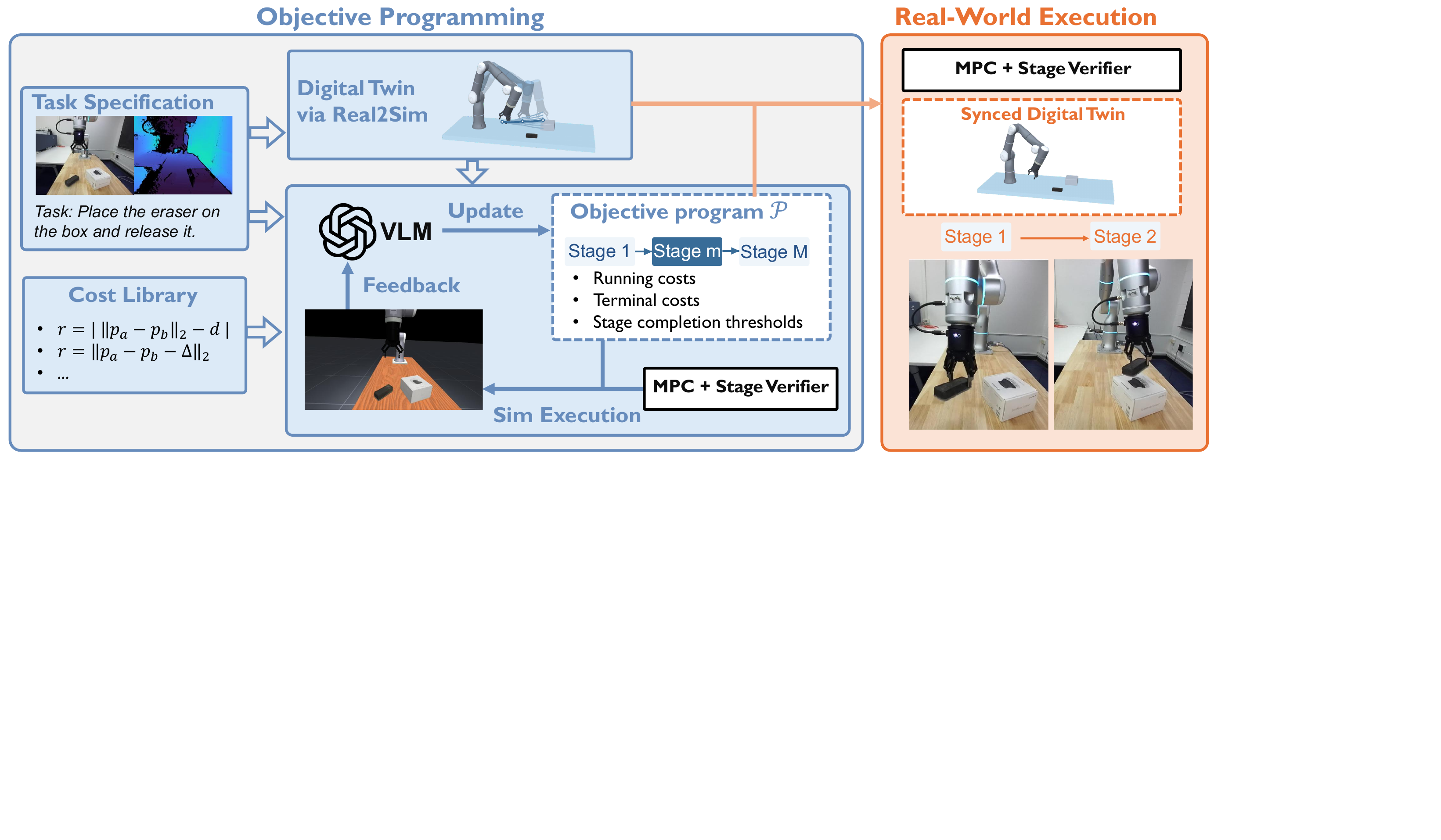}
\caption{\textbf{Method Overview.} Given the task specification in the form of a scene observation and a language description, we generate a digital twin through real-to-sim. VLM then generates objective programs given the task specification, the digital twin, and a predefined cost library. The objective program describes the running and
terminal objective terms for the MPC, on a potentially multi-stage task, along with completion conditions for every stage. The VLM iteratively improves the objective programs based on the rollouts in the digital twin, with an MPC that optimizes controls based on the objective program. A stage verifier decides when to advance to the next stage when all completion conditions are met. Finally, the object program is deployed in the real world with the same MPC while the digital twin is synced to the real world execution.}
\label{fig:system}
\end{figure*}

We introduce Objectives as Policies (\brand{}), a framework that expresses multi-stage tasks through an editable \emph{objective program} for a shared
sampling-based MPC controller. Each program combines
running and terminal objective terms from a fixed library into a sequence of stages,
along with completion conditions for every stage. A pretrained VLM
generates the program and iteratively refines the program through rollouts in a digital twin obtained from real-to-sim, before final closed-loop execution in the real world. We leverage two refinement strategies: the first uses the VLM to self-review against the task language instruction and the scene before any rollouts, and the second updates the objective program based on the statistics of a failed rollout, which can be done iteratively over many rounds. In simulation experiments, our method achieves 90\% cumulative success across four tasks with at most two feedback rounds and 97.5\% with up to five rounds. Real-world experiments on the same four tasks achieve an overall success rate of 80\% with deliberate scene perturbations.

\section{RELATED WORK}
\subsection{Foundation Models for Model-based Control}

Foundation models connect language instructions to task and motion planning through symbolic and geometric specifications. AutoTAMP translates instructions into temporal-logic specifications, while LLM$^3$ refines symbolic actions and continuous parameters using motion-planning failure feedback~\cite{chen2024autotamp,wang2024llm3}. VoxPoser and ReKep generate 3D value maps and relational keypoint constraints, respectively, for trajectory optimization~\cite{huang2023voxposer,huang2024rekep}.

Foundation models also specify objectives and guide action sampling for model-based control. Language to Rewards composes generic residual terms and adjusts their parameters and weights for MuJoCo MPC~\cite{yu2023l2r}. VLMPC guides action sampling and scores predicted video rollouts~\cite{zhao2024vlmpc}, while Hess et al. tune MPC objective weights using video feedback~\cite{hess2025sampling}. CoRAL supports multi-stage costs and revises objectives and physical-model parameters during execution~\cite{coral2026}. \brand{} combines geometric and contact-related residuals with stage-completion conditions in an editable objective program. The VLM refines this program through simulated execution before deployment, and real-world MPC replans controls under the resulting program.

\subsection{Sampling-Based MPC for Contact-Rich Manipulation}
Sampling-based MPC optimizes control sequences without differentiating
through contact dynamics, which is an attractive property for contact-rich manipulation. Sampling-based MPC methods mainly differ in how they weigh the sampled control rollouts to determine the optimal control. Predictive sampling selects the best control~\cite{howell2022mjpc}, model predictive path integral control (MPPI) uses a weighted average with exponential weights ~\cite{williams2017mppi}, and the cross-entropy method (CEM) uses a fitted Gaussian mean of the top samples~\cite{pinneri2020icem}. GPU-parallel physics rollouts support real-time planning for contact-rich tasks~\cite{li2025drop}.

\section{Method}
\label{sec:method}
Summarized by Fig.~\ref{fig:system}, our method first constructs a digital twin of the observed scene. The VLM uses the task instruction, scene observation, digital-twin information, and cost library to generate and self-review a multi-stage objective program. Each stage specifies running costs, terminal costs, and completion thresholds. MPC executes the task in simulation under these objectives, and failed executions provide feedback for program refinement. The resulting program then guides real-world execution with the synchronized digital twin. MPC replans controls from updated states, while the stage verifier advances the program when all completion conditions hold. In this section, we will first introduce the objective program in Sec.~\ref{sec:obj_program}, and discuss real-to-sim and MPC in Sec.~\ref{sec:r2s} and Sec.~\ref{sec:MPC}. We will then detail the objective functions and the cost library in Sec.~\ref{sec:obj}, and finish by describing the overall VLM workflow in Sec.~\ref{sec:VLM}.

\subsection{Objective Program}
\label{sec:obj_program}
An objective program for a multi-stage task specifies both the costs used for planning and the conditions used to advance stages. A stage $S = (\mathcal R, \mathcal T, w, \tau)$ is defined by the running cost terms $\mathcal R$, the terminal cost terms $\mathcal T$, the cost weights $w$, and the stage completion thresholds $\tau$. To reduce the large search space of possible cost terms, we pre-define a cost library $L$ of commonly used parametrized cost terms, and define the running and terminal cost terms $\mathcal R, \mathcal T$ as a combination of terms from the library with task-specific parameters, whose weighted sum according to weights $w$ becomes the objective function of the MPC controller. The conditions for completing stage $m$ are defined by the terminal terms $\mathcal T_m$ and the corresponding completion thresholds $\tau$. Then an objective program $\mathcal P=(S_1,\ldots,S_M)$ is defined by the costs and stage completion conditions of the $M$ stages. 

\subsection{Real-to-Sim}
\label{sec:r2s}
We build a digital twin of the scene to serve as the computation model of the MPC, both in simulation and real-world experiments, and as the rollout simulators in the objective program finetuning by VLM. For reconstructed objects, we obtain meshes from RGB observations
and masks using SAM~3D~\cite{sam3d2025}, then refine depth-based metric scale
and object pose with Any6D~\cite{any6d2025}. A VLM then estimates physical parameters of the objects, including density and friction coefficients, from object appearance~\cite {qwen3vl2025}. On hardware,
FoundationPose updates object poses from camera observations~\cite{wen2024foundationpose}. MuJoCo-warp is used as the computation model during MPC rollouts~\cite{todorov2012mujoco,mujocowarp2025}, and the original CPU-based MuJoCo is used as the rollout simulator since their contact dynamics are slightly more accurate than the MuJoCo-warp. While this simple pipeline works well for the tasks in our setup, in the future, we aim to perform explicit system identification of all shape, texture, and geometry parameters using differentiable simulation and rendering~\cite{oneshot} to minimize the sim-to-real gap.

\subsection{Model-Predictive Control}
\label{sec:MPC}

At each time step $t$, the MPC controller receives the updated scene states from the digital twin, plans a candidate control sequence $\mathbf U=(u_0,\ldots,u_{H-1})$ of horizon $H$ that optimizes an objective function $J^m_{\mathcal P}$ specified by the objective program and the current stage $m$, and applies the first $B$ controls. The controls are optimized by solving the following:
\begin{equation}
\begin{aligned}
    \mathbf U^\star &= \underset{\mathbf U \in \mathcal{U}}{\arg\min}\,J^m_{\mathcal P} (\mathbf U, \mathbf X) \\
    &\text{s.t.} \quad x_{i+1} = f_{sim}(x_i,u_i) \quad \forall x_i \in \mathbf  X.
\end{aligned}    
\end{equation}
Here $\mathbf X=(x_0,\ldots,x_H)$ is the state sequence generated by the controls according to the digital twin $f_{sim}$. $J^m_{\mathcal P}$ is the composite cost function generated according to the objective program generated by the VLM. Beyond the control limits, we use soft penalties to discourage constraint violations~\cite{nocedal2006numerical}. We adopt MPPI~\cite{williams2017mppi}, a sampling-based MPC controller, to solve this because it does not require gradients, which are hard to obtain for contact-rich manipulations, and benefits from GPU-based simulators, as all sample rollouts can be executed in parallel on the GPU.

At each MPC step, MPPI samples $K$ control sequences according to the Gaussian $\mathbf U^{(k)} \sim \mathcal{N}(\hat{\mathbf U},\Sigma)$, where $\hat{\mathbf U}$ is the nominal control sequence, and $\Sigma$ is a hyperparameter that we specify through pilot experiments. The nominal control sequence $\hat{\mathbf U}$ is obtained by copying the optimal solution from the last MPC update, cutting the first $B$ executed controls, and extending the sequence by repeating the last control $B$ times. For the very first MPC step, we use a zero mean. Each sample is then filtered by the control limits $\mathcal{U}$, and rolled out in parallel in the mujoco-warp simulator, generating the states $\mathbf X^{(k)}$. Here we abuse the notation a bit and include additional information, such as contact forces, in the states. Then MPPI calculates the optimal control as a weighted mean:
\begin{equation}
\label{eq:mppiupdate}
\begin{aligned}
\omega_k&=\frac{\exp(-(J^m_{\mathcal P}(\mathbf U^{(k)},\mathbf X^{(k)})-\rho)/\lambda)}
 {\sum_{k}\exp(-(J^m_{\mathcal P}(\mathbf U^{(k)},\mathbf X^{(k)})-\rho)/\lambda)},\\
\mathbf{U}^\star &=\sum_{k}\omega_k\mathbf U^{(k)}.
\end{aligned}
\end{equation}

Here, $\rho = \min_k J^m_{\mathcal{P}}(\mathbf{U}^{(k)},\mathbf{X}^{(k)})$ is the
smallest trajectory cost and $\lambda > 0$ is the temperature, where a lower
temperature concentrates the weights on the lower-cost sequences. We initialize $\lambda=1$ at the start of each stage and adapt it between MPC updates using the sum of unnormalized trajectory weights.

\subsection{Objective Function}
\label{sec:obj}
We provide the VLM with a fixed library of parameterized cost terms (Table~\ref{tab:objective-library}). The library restricts the objective search space while allowing the VLM to change term composition and parameters across stages. Each term defines an error residual $r$ that measures 1) geometric error, 2) violation of a grasp or a contact, or 3) gripper condition. Geometric errors include different measurements of errors in translational or rotational errors in desired pose or velocity. Grasp and contact costs use the contact indicators described in Table~\ref{tab:objective-library} to guide the MPC to make, break, and maintain contacts or grasps. The gripper condition is a hardware-specific term that specifies the desired gripper opening.

\begin{table*}[t]
\centering
\caption{Cost library: terms, input parameters, and residual definitions.
$r$ denotes residual error. }
\label{tab:objective-library}
\footnotesize
\setlength{\tabcolsep}{4pt}
\renewcommand{\arraystretch}{1.24}
\begin{tabularx}{\textwidth}{@{}>{\raggedright\arraybackslash}p{3.4cm}
  >{\raggedright\arraybackslash}p{4.7cm}
  >{\raggedright\arraybackslash}X@{}}
\toprule
\textbf{Term and Function Name} & \textbf{Input parameters} & \textbf{Residual definition} \\
\midrule
\textbf{Target distance}\newline
{\scriptsize\texttt{point\_point\_distance}} &
Points $a,b$, target distance $d$ &
$r=|\|p_a-p_b\|_2-d|$.\newline
$p_a,p_b$: world-frame positions of points $a,b$;
$d$: target distance. \\
\textbf{Relative position}\newline
{\scriptsize\texttt{relative\_position}} &
Points $a,b$, desired world-frame offset $\Delta$ &
$r=\|p_a-p_b-\Delta\|_2$.\newline
$p_a,p_b$: world-frame positions of points $a,b$;
$\Delta$: desired world-frame offset. \\
\textbf{Axis alignment}\newline
{\scriptsize\texttt{axis\_parallel}} &
Axes $a,b$, alignment direction $\eta\in\{-1,1\}$ &
$r=\arccos(\eta v_a^\top v_b)$.\newline
$v_a,v_b$: world-frame unit axes; $\eta=1$: parallel,
$\eta=-1$: antiparallel. We clip the dot product to $[-1,1]$.
Angles are in radians. \\
\textbf{Target angle}\newline
{\scriptsize\texttt{axis\_angle}} &
Axes $a,b$, target angle $\theta^\star$ &
$r=|\arccos(v_a^\top v_b)-\theta^\star|$.\newline
$v_a,v_b$: world-frame unit axes; $\theta^\star$: target angle in radians.
We clip the dot product to $[-1,1]$. \\
\textbf{Directional gap}\newline
{\scriptsize\texttt{signed\_axis\_gap}} &
Points $a,b$, unit axis $n$, direction $\sigma\in\{-1,1\}$,
minimum gap $m$ &
$r=[m-\sigma n^\top(p_b-p_a)]_+$.\newline
$p_a,p_b$: world-frame positions of points $a,b$; $n$: world-frame unit axis;
$\sigma$: direction sign; $m$: minimum gap; $[z]_+=\max(0,z)$. \\
\textbf{Minimum rotation}\newline
{\scriptsize\texttt{rotation\_progress}} &
Object $o$, initial orientation $R_0$, unit axis $n$ in the initial
frame, minimum angle $\theta_{\min}$ &
$\alpha=(R_0n)^\top\operatorname{rotvec}(R_oR_0^\top)$,\newline
$r=[\theta_{\min}-\alpha]_+$.\newline
$R_o,R_0$: current and episode-start local-to-world orientations of object $o$;
$n$: initial-frame unit axis; $\operatorname{rotvec}$: principal rotation vector.
$\alpha$: signed rotation along $R_0n$; $\theta_{\min}$: minimum angle in radians;
$[z]_+=\max(0,z)$. \\
\textbf{Point-to-line distance} \newline {\scriptsize\texttt{point\_on\_line}} & Points $a,b$, unit line direction $n$ & $r=\|(p_a-p_b)\times n\|_2$.\newline $p_a,p_b$: world-frame point positions; $n$: world-frame unit direction of the line through $p_b$. The term allows movement along the line. \\ \textbf{Point in region} \newline {\scriptsize\texttt{in\_region}} & Point $a$, world-axis-aligned box with center $c$ and half-sizes $h$ & $r=\displaystyle\max_{k\in\{x,y,z\}}\frac{[|p_{a,k}-c_k|-h_k]_+}{h_k}$.\newline $p_a,c$: world-frame point and region-center positions; $h_k>0$: region half-size along world axis $k$; $[z]_+=\max(0,z)$. The term measures normalized boundary violation for point $a$. \\ \midrule \textbf{Tool-mediated contact} \newline
{\scriptsize\texttt{tool\_mediation}} &
Tool $o$, target $q$, contact requirement $\rho\in\{0,1\}$ &
$\displaystyle\mathbf r=\begin{bmatrix}
M_{\mathrm{robot},q}\\ \rho(1-M_{o,q})
\end{bmatrix}$.\newline
$M_{\mathrm{robot},q},M_{o,q}\in\{0,1\}$ indicate robot--target and
tool--target contact anywhere in the predicted trajectory.
Here $o$ is the tool, $q$ the target, and $\rho=1$ requires tool--target contact. \\
\textbf{Object held} \newline
{\scriptsize\texttt{object\_held}} &
Object $o$ & $r_{o,h}=1-\chi_{o,h}$.\newline
$\chi_{o,h}=1$ when both fingers contact object $o$ at predicted physics
step $h$, and zero otherwise. \\
\textbf{Object not held} \newline
{\scriptsize\texttt{not\_held}} &
Object $o$ & $r_{o,h}=\chi_{o,h}$.\newline\\
\midrule
\textbf{Gripper command} \newline
{\scriptsize\texttt{gripper\_command}} &
Target opening width $g^\star$ ($0$: close, $W$: open) &
$g(u_g)=\frac{W}{2}(1-\operatorname{clip}(u_g,-1,1))$,\newline
$r=|g(u_g)-g^\star|$.\newline
$u_g$: normalized gripper effort command ($+1$: close, $-1$: open);
$g(u_g)$: equivalent opening width; $W=85\,\mathrm{mm}$: fully open width;
$g^\star$: target width. \\
\bottomrule
\end{tabularx}
\end{table*}

Following residual-based cost design~\cite{howell2022mjpc}, we then map these residuals to running and terminal cost terms used by the MPC. The running costs penalize the residuals, while the terminal costs allow a dead band with zero cost, according to the stage termination thresholds. Each running or terminal term will be accompanied by a scale $s_i>0$ with the same units as $r_i$, making $r_i/s_i$ dimensionless. The running cost is then defined as:
\begin{equation}
    c_i^{\mathrm R}(x,u)=\sqrt{1+(r_i(x,u)/s_i)^2}-1,
\end{equation}
This residual-to-cost mapping is approximately quadratic near zero and approximately linear for large residuals. The terminal cost is
\begin{equation}
\label{eq:stagecost}
c_i^{T}(x,u,\tau_i)= \left(\frac{[r_i(x,u)-\tau_i]_+}{s_i}\right)^2,
\end{equation}
where $\tau_i$ is the threshold for term $i$, and the term incurs zero cost when $r_i\leq\tau_i$.

The objective function of the MPC $J_{\mathcal P}^{m}$ is then defined as:

\begin{equation}
\label{eq:objective}
\begin{aligned}
J_{\mathcal P}^m&(\mathbf X,\mathbf U)=\frac{1}{H}\sum_{c^R_i \in \mathcal{R}_m} w_i\sum _tc^R_i(x_t,u_t)\\ &+\sum_{c^T_j \in \mathcal{T}_m} w_{j}c^T_j(x_{H}, u_{H-1}, \tau_j)
+C_{\mathrm{physics}}(\mathbf X,\mathbf U).
\end{aligned}
\end{equation}

Here, the first two terms are running and terminal terms in $\mathcal R_{m}$ and $\mathcal{T}_m$, that are a weighted sum of terms from the objective library $L$ weighted by $w$. The additional physics-based cost $C_{\mathrm{physics}}$ includes terms that penalize control variation, joint velocities, large contact forces, and pose changes of movable objects not specified by the VLM. We include these terms in every stage to discourage physically undesirable behavior.

\paragraph{Stage Verifier}
In order to verify the completion threshold in the terminal, a separate verifier at each time step compares each terminal residual with its corresponding threshold $\tau$ and only advances to the next stage when all terminal conditions of that stage are satisfied.

\paragraph{Example Cost Function Generated by VLM: carrying the eraser}
A carry stage from the pick-and-place task needs to move the eraser's center
$0.0465\,\mathrm m$ vertically above the center of the box's top face
while maintaining the grasp and aligning their top-face normals. The generated MPC running costs include:
\begin{itemize}
    \item Target position with \texttt{relative\_position}.
    \item Target axis alignment  with \texttt{axis\_parallel}.
    \item Encourage a closing command with \texttt{gripper\_command}.
    \item Gripper maintains contact with object with \texttt{object\_held}.
\end{itemize}
The terminal costs are:
\begin{itemize}
    \item The gripper maintains contact with the object with \texttt{object\_held}.
    \item Target position with \texttt{relative\_position} with a threshold of 10\,mm.
\end{itemize}

\subsection{Iterative Objective Program Refinement via VLM}
\label{sec:VLM}
For the VLM to generate a draft $\mathcal P^{\mathrm{draft}}$, we provide it with a scene observation $o$ and a natural language instruction $\ell$ of the intended task, along with the information in the digital twin, such as all object 3D locations, and the cost library in
Table~\ref{tab:objective-library}. It then self-reviews the draft once to obtain $\mathcal P^{(0)}$. Section~\ref{sec:results} evaluates self-review with and without execution feedback. The objective programs are then iteratively refined by deploying the MPC in the digital twin, obtaining the statistics of the rollout, and updating the program based on this feedback, generating a sequence of programs $\mathcal P^{(0)},\mathcal P^{(1)},\mathcal P^{(2)}, \cdots$. Finally, physical execution occurs after success is achieved in simulation or the refinement budget is met, with the latest program.

In addition to all the inputs mentioned above, the VLM prompt also asks the model to divide the task into stages, with running costs that guide motion and terminal conditions that specify when each stage is complete. It also asks for a brief explanation of each stage's objectives and how that stage could fail during execution. Finally, it specifies the required output format of the program in JSON. The objective program is a JSON object with \texttt{program} and
\texttt{reasoning} fields. The \texttt{program} field contains an ordered
\texttt{stages} list. Each stage specifies its \texttt{name}, its
\texttt{running} and \texttt{terminal} cost terms, and the objects
it intends to move (\texttt{movable\_objects}). For each cost term, the VLM gives the term name (\texttt{type}) and
the input parameters listed in
Table~\ref{tab:objective-library}. For example, a target-distance term
specifies two points (in space or on an object) and a target distance.
During self-review, the model checks whether each later stage can start from
the preceding stage's end state. It also checks for conflicting objectives
and infeasible grasp or contact requirements, then revises the draft
where needed. 

After the initial simulated execution of $\mathcal P^{(0)}$, if not successful, we return the program and a record of the execution to the VLM. The feedback prompt includes the original inputs and design rules, plus $n$ measured state records uniformly subsampled across the entire execution, including the first and last available records. We set $n=6$ in our experiments. These measurements include object and gripper poses, the active stage, gripper width, contact measurements, residual values, and whether the controller reports that an object is held. We also list the stages reached during execution and the last recorded measurements for each stage's completion conditions. The VLM uses these observations to revise the objective program.

\section{EXPERIMENTS}
\label{sec:setup}
Our evaluation aims to examine whether the generated
objective programs are effective in driving the MPC controller to complete contact-rich manipulation tasks and how
performance changes with the objective refinement. We will
also investigate the effects of our algorithmic innovations
compared to previous methods.

\begin{figure*}[t]
\centering
\includegraphics[width=170mm]{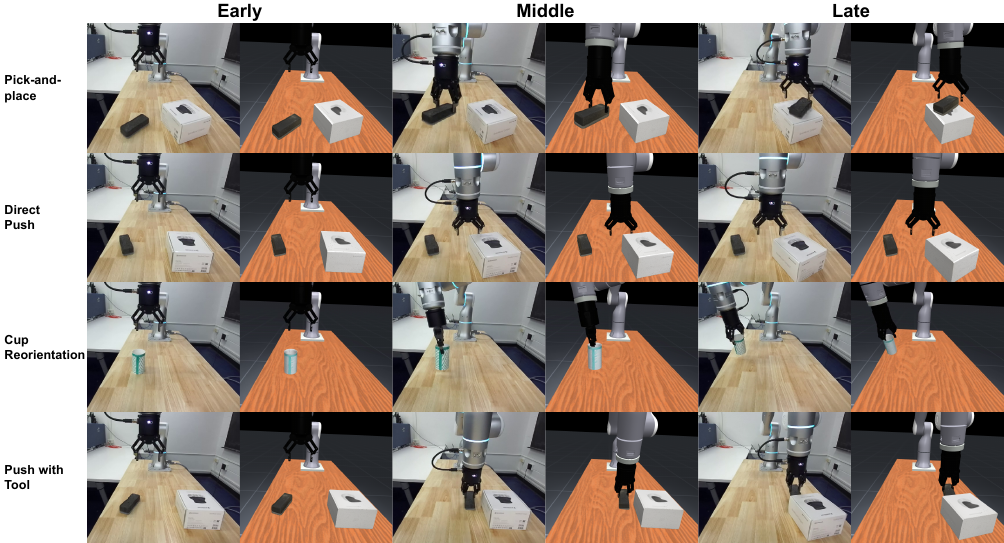}
\caption{\textbf{Real-world execution across four tasks.} Each pair shows a real-camera frame (left) and its MuJoCo reconstruction (right) from the same run's recorded robot and object states. } 
\label{fig:camera-setup}
\end{figure*}

\subsection{Experimental Setup}
\label{sec:criterion}
\label{sec:budget}
Our platform uses a seven-degree-of-freedom Flexiv Rizon 4s arm with a GN01
gripper and a ZED 2i RGB-D camera (Fig.~\ref{fig:camera-setup}). We calibrate the camera pose relative to
the robot base and reproduce it in MuJoCo. We evaluate on four tabletop manipulation tasks:
\begin{itemize}
\item \textbf{Pick-and-place:} Place the blackboard eraser on top of the
white product box and release it.
\item \textbf{Cup reorientation:} Grasp the cup and tilt it by at least
$20^\circ$ toward robot-base $+X$ while keeping it held.
\item \textbf{Direct push:} Push the white box at least 60\,mm along
robot-base $+X$ through direct robot-box contact.
\item \textbf{Push with Tool:} Use the blackboard eraser to push the white
product box at least 60\,mm along robot-base $+X$ while keeping the eraser held.
\end{itemize}
We evaluate task success using fixed, task-specific criteria independent of the generated stage completion thresholds. We use simulator ground-truth states in simulation and measured states on hardware. The stage verifier controls program progression, while the independent task criteria determine the reported success rates.

For all experiments, we use GPT-5.6-sol with \texttt{high} reasoning effort for generation and
refinement~\cite{openai2026gpt56}, without updating model weights. All OaP experiments use the same MPC settings. Each update uses $N=1000$ optimization rollouts over $H=600$ physics steps at 2\,ms and applies the first $B=40$ control inputs before replanning. Each execution permits at most 250 MPC updates and 20\,s of commanded motion, with a 25\,N closing-force
limit. 

\subsection{Simulated Experiments}\label{sec:results}
We repeat VLM generation and MPC execution 10 times with the same simulated scene for each task, and report the results in Table II. We show the success rates for both with and without self-review, and for increasing feedback rounds up to five rounds. We can see that self-review raises first-execution success from 20/40 to 26/40 across the four tasks(Table II) with zero feedback rounds. The difference narrows as more feedback rounds are used. We hypothesize that during self-review, a VLM can identify inconsistencies visible in the program, such as whether a subtask can start from the state left by the preceding subtask. However, the self-review round cannot foresee failures that only arise during physical execution, such as slipping grasp. These failures can be addressed only after they are observed and provided to the VLM through execution feedback. This may explain why self-review provides the largest improvements during the first execution and why its relative benefit decreases over subsequent feedback rounds. Self-review can potentially add intermediate objectives beyond the task requirement that can effectively improve its success rate.

With at most five feedback rounds, success reaches 95\% without self-review and 97.5\% with self-review. One pick-and-place trial illustrates how feedback improved an objective program. In the first feedback revision, the VLM retained an \texttt{object\_held} running cost in the last stage. As a result, the eraser remained in place at the end of execution, preventing it from being placed in the target location. In the second feedback round, the VLM removed this cost and added another settling stage,  in which the next execution successfully placed and released the eraser. 

\begin{table*}[t]
\centering
\caption{Cumulative simulation success across feedback budgets.
Each task has ten trials per condition. Earlier successes carry forward, and each feedback round permits at most one additional execution.}
\label{tab:ablation-design}
\label{tab:feedback-rounds}
\setlength{\tabcolsep}{4pt}
\small
\begin{tabular*}{\textwidth}{@{\extracolsep{\fill}}l*{12}{c}@{}}
\toprule
& \multicolumn{6}{c}{Without Self-review} & \multicolumn{6}{c}{Self-review}\\
\cmidrule(lr){2-7}\cmidrule(lr){8-13}
Task / feedback rounds & 0 & 1 & 2 & 3 & 4 & 5 & 0 & 1 & 2 & 3 & 4 & 5\\
\midrule
Pick-and-place & 6 & 9 & 10 & 10 & 10 & 10 & 9 & 10 & 10 & 10 & 10 & 10\\
Cup Reorientation& 4 & 7 & 8 & 8 & 9 & 9 & 4 & 6 & 7 & 7 & 8 & 9\\
Direct Push & 4 & 7 & 8 & 9 & 9 & 9 & 7 & 10 & 10 & 10 & 10 & 10\\
Push with Tool & 6 & 8 & 10 & 10 & 10 & 10 & 6 & 8 & 9 & 10 & 10 & 10\\
\midrule
Total & 20/40 & 31/40 & 36/40 & 37/40 & 38/40 & 38/40 & 26/40 & 34/40 & 36/40 & 37/40 & 38/40 & 39/40\\
Success rate & 50\% & 77.5\% & 90\% & 92.5\% & 95\% & 95\% & 65\% & 85\% & 90\% & 92.5\% & 95\% & 97.5\%\\
\bottomrule
\end{tabular*}
\end{table*}

\subsection{Comparison with Baselines and Ablations in Simulated Experiments}
We compare \brand{} with Language to Rewards (L2R)~\cite{yu2023l2r},
ReKep~\cite{huang2024rekep}, OpenVLA-OFT~\cite{kim2025oft}, and
$\pi_{0.5}$~\cite{physicalintelligence2025pi05}, and CoRAL~\cite{coral2026}. For the VLA baselines, we retain the robot and camera configuration of the released LIBERO evaluation setups to avoid introducing a new embodiment or camera arrangement. This includes the Franka Emika Panda, PandaGripper, and operational-space pose controller, with third-person and wrist-camera inputs~\cite{kim2025oft,physicalintelligence2025pi05}.
We preserve the third-person camera pose and wrist-camera mounting relative to the hand, together with the upstream image preprocessing. The four task goals and objects come from our manipulation scenarios. Thus, these experiments evaluate transfer to our custom task scenes using the released checkpoints without task-specific fine-tuning. While the authors of CoRAL did not fully open-source their code, we follow their published method as closely as possible. For comparison with OaP, we use the same GPT-5.6-sol model with high reasoning effort, providing scene observations alongside language instructions.

We report the results in Table~\ref{tab:external-baselines}, using our method with only 2 rounds of feedback. While the VLAs struggle with these tasks, methods that leverage model-based planning perform better. CoRAL is the closest to ours, but comes at the cost of very frequent LLM updates (1 Hz). This highlights the importance of having a compact objective function library, as our method already achieves a better success rate with only one feedback round. Our ablation also shows lower success without the library: 20\% versus 50\% without self-review and 35\% versus 65\% with self-review (Table III).

\begin{table*}[t]
\centering
\caption{Comparison with baselines and ablations in simulation.}
\label{tab:external-baselines}
\setlength{\tabcolsep}{5pt} \footnotesize \renewcommand{\arraystretch}{0.95} \begin{tabular*}{\textwidth}{@{\extracolsep{\fill}}lccccc@{}}
\toprule
Method & Pick-and-place & Cup Reorientation & Direct Push & Push with Tool & Overall\\
\midrule
\multicolumn{6}{l}{\textit{Baselines }}\\ L2R~\cite{yu2023l2r} & 2/10 & 3/10 & 4/10 & 2/10 & 11/40 (27.5\%)\\
ReKep~\cite{huang2024rekep} & 9/10 & 6/10 & 0/10 & 2/10 & 17/40 (42.5\%)\\
CoRAL (adapted)~\cite{coral2026} & 7/10 & 6/10 & 10/10 & 8/10 & 31/40 (77.5\%)\\
OpenVLA-OFT~\cite{kim2025oft} & 0/10 & 0/10 & 0/10 & 0/10 & 0/40 (0\%)\\
$\pi_{0.5}$~\cite{physicalintelligence2025pi05} & 10/10 & 0/10 & 0/10 & 0/10 & 10/40 (25\%)\\
\midrule
\multicolumn{6}{l}{\textit{OaP ablations (no feedback)}}\\ w/o library, w/o Self-review & 1/10 & 1/10 & 3/10 & 3/10 & 8/40 (20\%)\\ w/o library, with Self-review & 5/10 & 3/10 & 3/10 & 3/10 & 14/40 (35\%)\\ with library, w/o Self-review & 6/10 & 4/10 & 4/10 & 6/10 & 20/40 (50\%)\\ with library, with Self-review & 9/10 & 4/10 & 7/10 & 6/10 & 26/40 (65\%)\\ \midrule \multicolumn{6}{l}{\textit{OaP with $2$ feedback rounds}}\\ \brand{} (library + Self-review) & 10/10 & 7/10 & 10/10 & 9/10 & 36/40 (90\%)\\
\bottomrule
\end{tabular*} \end{table*}

\subsection{Real-World Experiments}
For each task, we select five objective programs that succeed in simulation and execute each once under both physical scene configurations in Table~\ref{tab:real-world}. We keep the programs fixed and perturb object locations relative to the refinement scene. These programs succeed in 32/40 hardware executions, measuring transfer from successful simulation programs to physical scenes with changed object locations. Most failures in pick-and-place, cup reorientation, and push with the tool were from poor grasp quality. We hypothesize the VLM estimated friction coefficients used by the simulator are far from the actual value of the real objects. The friction mismatch is most evident with the 3D-printed cup, where MPC predicts a stable grasp in simulation but the same motion slips on the real robot. In addition, the simulator treats every object as rigid, whereas real objects deform under contact. Because the gripper applies a fixed closing force, objects of different stiffness respond differently to the same grasp, which the simulator does not capture. This shows that the accuracy of the digital twin will be a key bottleneck.

\begin{table}[t]
\centering
\caption{Real-world success of simulation-successful programs. Each program runs once per scene configuration.}
\label{tab:real-world}
\setlength{\tabcolsep}{3pt}
\small
\begin{tabular}{lccc}
\toprule
Task & Small Perturb. & Large Perturb. & Overall\\
\midrule
Pick-and-place & 4/5 & 4/5 & 8/10 (80\%)\\
Cup Reorientation & 4/5 & 3/5 & 7/10 (70\%)\\
Direct Push & 4/5 & 4/5 & 8/10 (80\%)\\
Push with Tool & 5/5 & 4/5 & 9/10 (90\%)\\
\midrule
Total & 17/20 & 15/20 & 32/40 (80\%)\\
\bottomrule
\end{tabular}
\end{table}

\section{DISCUSSION}
Objective programs provide an explicit interface for specifying, executing, and revising manipulation tasks. OaP uses simulated execution feedback to revise these programs before deployment, then uses a shared MPC controller to execute them from updated physical states.

\subsection{Limitations}
The scope of \brand{} depends on its residual library and physical model. The fixed library defines the objectives that the VLM can express.
Real-world execution also depends on the accuracy of scene reconstruction and contact dynamics, particularly under occlusion and grasp slip.

The controller also lacks explicit recovery from failures after a stage has completed. The objective sequence generated by the VLM follows a linear, forward-only plan: once the verifier advances to the next stage, the controller cannot return to an earlier one if its completion condition later becomes invalid. Some failure recovery may still be observable when objective terms from a completed stage are carried into subsequent stages, encouraging the controller to re-establish the corresponding conditions. However, this behavior is incidental rather than an explicit recovery mechanism. Explicit stage recovery is a direction for future work.
\section{CONCLUSION}
Objectives as Policies uses an executable objective program to connect
VLM task specification with closed-loop manipulation. The program expresses task goals as a multistage cost function, which guides a shared
sampling-based MPC controller as it optimizes control sequences using
physics rollouts from the current state. Because the goals remain explicit in the program, the VLM can reconsider them through self-review before execution and execution feedback afterward. The objective program thus provides a common interface for task specification, execution, and revision without task-specific policy training. 

We expect the framework to keep improving manipulation as VLMs get better. For future work, we aim to include online real-to-sim and real-world execution in the VLM feedback loop, where the digital twin's parameters can be estimated on the fly, and VLM can improve cost programs based on physical execution, and even tune the MPC hyperparameters. We would also like to compare against agentic frameworks for robot manipulation.

\bibliographystyle{IEEEtran}
\bibliography{refs,refs_YZ}
\end{document}